\def\year{2020}\relax
\documentclass[letterpaper]{article}
\usepackage{aaai20}
\usepackage{times}
\usepackage{helvet}
\usepackage{courier}
\usepackage{url}
\usepackage{graphicx}
\newcommand{\citet}[1]{\citeauthor{#1} \shortcite{#1}}
\newcommand{\citep}{\cite}

\usepackage{multirow}
\usepackage{booktabs}
\usepackage{acronym}
\usepackage{color}
\usepackage{amssymb}

\usepackage{subcaption}
\usepackage[inline]{enumitem}
\usepackage{amsmath}
\usepackage{array}

\usepackage[skip=0pt]{caption}

\usepackage[subtle, wordspacing=normal, tracking=normal, bibnotes, charwidths=normal, leading, indent, paragraphs=normal, mathspacing=normal, bibliography=normal]{savetrees}

\acrodef{KGCS}{Knowledge Grounded Conversational System}
\acrodef{BBC}{Background Based Conversation}
\acrodef{KS}{Knowledge Selection}
\acrodef{GKS}{Global Knowledge Selection}
\acrodef{LKS}{Local Knowledge Selection}
\acrodef{GLKS}{Global-to-Local Knowledge Selection}
\acrodef{DS}{Distant Supervision}
\acrodef{MLE}{Maximum Likelihood Estimation}
\acrodef{MCE}{Maximum Causal Entropy}
\acrodef{BiDAF}{Bi-directional Attention Flow}
\acrodef{GTTP}{Get To The Point}
\acrodef{MRC}{Machine Reading Comprehension}
\acrodef{S2SA}{Sequence-to-Sequence with Attention}

\DeclareMathOperator{\softmax}{softmax}
\DeclareMathOperator{\BiGRU}{BiGRU}
\DeclareMathOperator{\GRU}{GRU}
\DeclareMathOperator{\attention}{attention}
\DeclareMathOperator{\Jaccard}{Jaccard}

\title{Thinking Globally, Acting Locally: Distantly Supervised Global-to-Local Knowledge Selection for Background Based Conversation}
\author{Pengjie Ren\textsuperscript{\rm 1}, Zhumin Chen\textsuperscript{\rm 2}, Christof Monz\textsuperscript{\rm 1}, Jun Ma\textsuperscript{\rm 2}, Maarten de Rijke\textsuperscript{\rm 1}\\
\textsuperscript{\rm 1} University of Amsterdam, Amsterdam, The Netherlands\\
\textsuperscript{\rm 2} Shandong University, Jinan, China\\
\{p.ren, c.monz, derijke\}@uva.nl, \{chenzhumin, majun\}@sdu.edu.cn
}

\begin{document}

\maketitle

\begin{abstract}
\acp{BBC} have been introduced to help conversational systems avoid generating overly generic responses.
In a \ac{BBC}, the conversation is grounded in a knowledge source.
A key challenge in \acp{BBC} is \ac{KS}: given a conversational context, try to find the appropriate background knowledge (a text fragment containing related facts or comments, etc.) based on which to generate the next response.
Previous work addresses \ac{KS} by employing attention and/or pointer mechanisms.
These mechanisms use a \emph{local} perspective, i.e., they select a token at a time based solely on the current decoding state.
We argue for the adoption of a \emph{global} perspective, i.e., pre-selecting some text fragments from the background knowledge that could help determine the topic of the next response.
We enhance \ac{KS} in \acp{BBC} by introducing a \ac{GLKS} mechanism.
Given a conversational context and background knowledge, we first learn a topic transition vector to encode the most likely text fragments to be used in the next response, which is then used to guide the local \ac{KS} at each decoding timestamp.
In order to effectively learn the topic transition vector, we propose a distantly supervised learning schema.
Experimental results show that the \ac{GLKS} model significantly outperforms state-of-the-art methods in terms of both automatic and human evaluation.
More importantly, \ac{GLKS} achieves this without requiring any extra annotations, which demonstrates its high degree of scalability.
\end{abstract}


\section{Introduction}
Non-task-oriented conversational systems (a.k.a., chatbots) aim to engage users in conversations for entertainment \citep{ijcai2018-778} or to provide valuable information~\citep{zhou2018dataset}.
Sequence-to-sequence models are an effective framework that is commonly adopted in this field.
However, a problem of vanilla sequence-to-sequence based methods is that they tend to generate generic and non-informative responses with bland and deficient responses~\citep{chen2017survey}.
Various methods have been proposed to alleviate this issue, such as adjusting objective functions~\citep{li2016diversity,NIPS2018_7452,jiang-2019-improving} or incorporating personal profiles~\citep{zhang2018personalizing}.

\acfp{BBC} have demonstrated a potential for generating more informative responses \citep{zhou2018dataset}. 
Given some background knowledge (e.g., an article in the form of free text) and a conversation, the \ac{BBC} task is to generate responses by referring to the background knowledge and considering the dialogue history context at the same time.
A key challenge in \acp{BBC} is \acfi{KS}, which is the task of finding the appropriate background knowledge (e.g., a text fragment about a movie plot) based on which the next response is to be generated.

Existing methods for \acp{BBC} can be grouped into two categories: extraction-based methods and generation-based methods.
The former addresses \ac{KS} by learning two pointers to extract spans from the background material as responses, and outperforms generation-based methods in finding knowledge~\citep{moghe2018towards}.
However, there are two major issues with extraction-based methods.
First, in most cases the generated responses are not natural due to their extractive nature.
Second, unlike, e.g., \ac{MRC}, in \acp{BBC} there is no notion of standard answer.
For example, extraction-based methods cannot handle greetings in chitchats.
%
\begin{figure}[t]
\begin{subfigure}{\columnwidth}
  \centering
  \includegraphics[width=0.95\columnwidth]{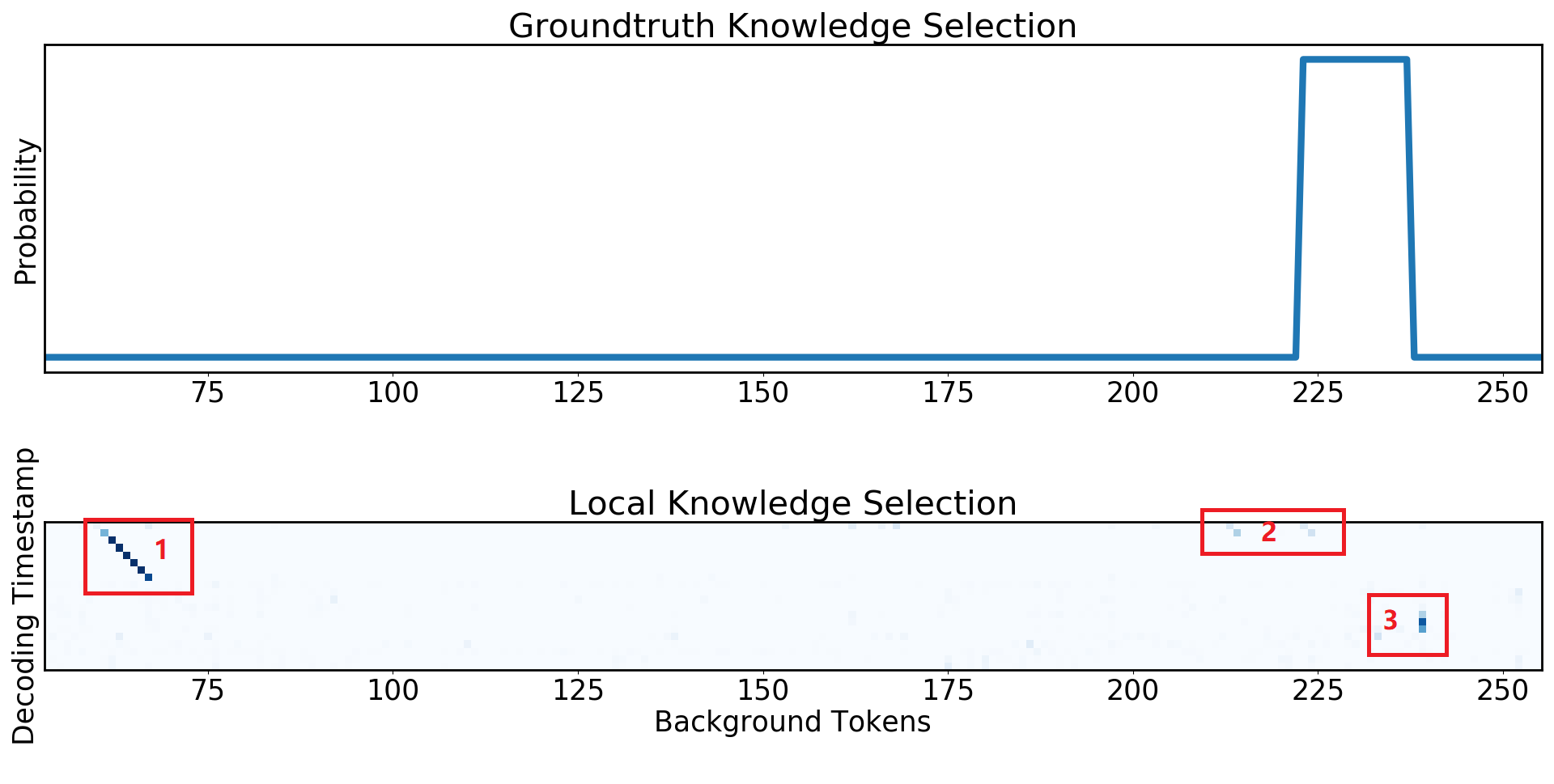}  
\end{subfigure}
\caption{Visualization of local knowledge selection. The X-axes represent the background tokens; the top Y-axis represents \ac{KS} probabilities and the spike indicates the ground truth \ac{KS}; the bottom Y-axis represents the decoding timestamp and darker blue means larger \ac{KS} probabilities.}
\label{fig:01}
\end{figure}

Today's generation-based methods perform \ac{KS} with a local perspective, i.e., by selecting one token at a time based solely on the current decoding state.
This is problematic because they lack the guidance that a more global perspective would offer.
In Figure~\ref{fig:01}, we visualize the \ac{KS} of a state-of-the-art model, an improved \ac{GTTP}, which achieves competitive performance on this task.
The top figure corresponds to the ground truth \ac{KS} annotations; the lower figure shows the \ac{KS} probabilities of \ac{GTTP} at each decoding timestamp.
\ac{GTTP} settles on two background areas (red box 1 and 2) at first in a sign of hesitation.

However, due to the lack of a global perspective, it chooses the wrong one (red box 1).
And it is too late when \ac{GTTP} realizes this and tries to correct its mistakes (red box 3).
In this paper, we propose to address this issue and enhance \ac{KS} for generation-based methods by introducing a \acfi{GLKS} mechanism.
The general idea is that we learn a ``topic transition vector'' with a \ac{GKS} module beforehand, which sets the tone for the next response and encodes the general meaning of the most likely used background knowledge.
The ``topic transition vector'' is then used to guide the \ac{LKS} at each decoding timestamp to avoid situations like the one in Figure~\ref{fig:01}.

As in existing work, we train \ac{LKS} with the \ac{MLE} loss.
However, \ac{MLE} is not effective enough to supervise the learning of \ac{GKS} because it only provides token-wise supervision.
To this end, we propose a distantly supervised learning schema where we use the Jaccard similarity between the ground truth responses and the background knowledge as an extra signal to train \ac{GKS}.
All parameters are learned by a linear combination of the global \ac{DS} and local \ac{MLE} in an end-to-end back-propagation training paradigm.

Several recent studies try to improve the \ac{KS} of generation-based methods.
\citet{chuanmeng2019} introduce a reference decoder that learns to directly select a semantic unit (e.g., a span containing complete semantic information) from the background, besides generating the response token by token.
\citet{liu2019knowledge} fuses two types of knowledge, triples from a structured knowledge graph and texts from unstructured background material, for better \ac{KS}.
Although they achieve promising improvements, they all have obvious limitations.
\citet{chuanmeng2019}'s work needs boundary annotations of semantic units in both backgrounds and responses to enable supervised training.
To be able to put \citet{liu2019knowledge}'s model to work, the authors prepare a structured knowledge source and manually ground unstructured background to it beforehand.
To show the effectiveness of \ac{GLKS}, we carry out experiments on the same datasets as \citet{chuanmeng2019} and \citet{liu2019knowledge}.
Our proposed \ac{GLKS} model significantly outperforms their models as well as other state-of-the-art methods in terms of both automatic and human evaluation.
\ac{GLKS} is able to generate natural responses, yielding better \ac{KS}, while
requiring minimum efforts (in terms of human annotations), which means it exhibits better scalability.

Our contributions are summarized as follows:
\begin{itemize}[leftmargin=*,nosep]
\item We propose a novel neural architecture with a \acf{GLKS} mechanism for \acp{BBC} that can generate more appropriate responses while retaining fluency. 
\item We devise an effective combined global (\ac{DS}) and local (\ac{MLE}) learning schema for \ac{GLKS} without using extra annotations.
\item Experiments show that \ac{GLKS} outperforms state-of-the-art models by a large margin in terms of both automatic and human evaluation.
\end{itemize}


\section{Related Work}

\subsection{Open-domain Conversation}
Sequence-to-sequence modeling for open-domain conversations has been studied for years \citep{shang2015neural}.
Previous studies have proposed various variants on different conversational tasks \citep{lowe2015ubuntu,serban2016building} and have shown the superiority of sequence-to-sequence conversation modeling when compared to IR or template-based methods, especially in generating fluent responses. 
However, many challenges remain.
Response informativeness is especially important; conversations become dull and less attractive due to too many generic responses such as ``I don't know'' and ``I am sorry'' \citep{vougiouklis2016neural,he2017generating}.
A number of studies address this issue by promoting response diversity.
They either propose new losses \citep{li2016diversity,zhao2017learning,jiang-2019-improving} or introduce new learning schemas \citep{NIPS2018_7452}.
Another strategy is to incorporate latent topic information \citep{xing2017topic} or leverage external knowledge~\citep{ghazvininejad2018knowledge,liu2018knowledge,zhou2018commonsense,young2018augmenting}. 

\subsection{Background Based Conversation}
\acfp{BBC} have shown promising results in improving response informativeness \citep{zhou2018dataset,dinan2018wizard,qin-etal-2019-conversing}.
Work on \acp{BBC} can be grouped into extraction-based and generation-based methods.

\begin{figure*}[ht]
\centering
\includegraphics[width=1.9\columnwidth]{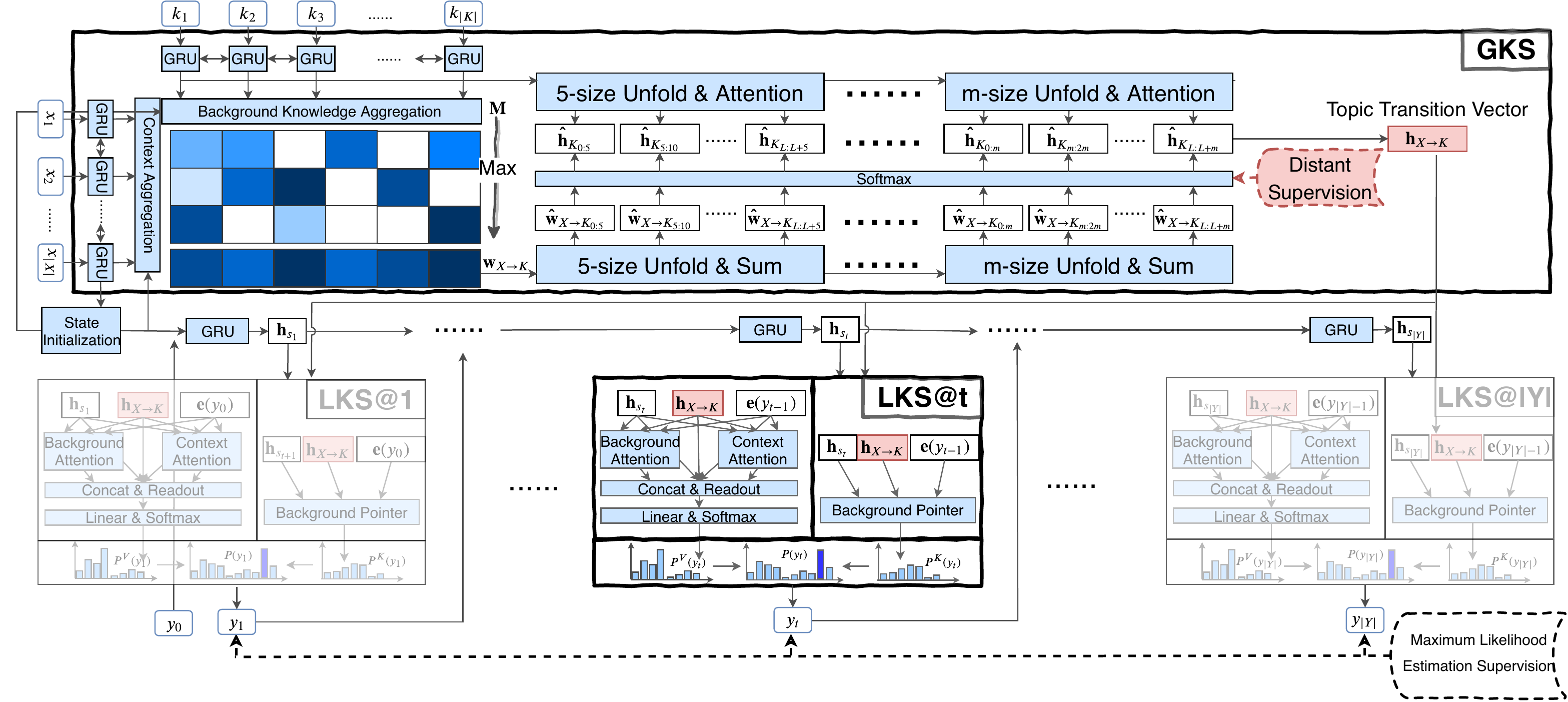}
\caption{Overview of \acf{GLKS}.}
\label{fig:02}
\end{figure*}

Extraction-based methods grew out of work on \acf{MRC}, where a span is extracted from the background as response to a question~\citep{seo2016bidirectional}.
Extraction-based methods are good at locating the right background knowledge but because they are designed for \ac{MRC} tasks, where user utterances are mostly simple questions that can be answered by a span, they are not suitable for \acp{BBC} \citep{moghe2018towards}.
The extracted spans are not natural as conversational responses, and in many cases there are no standard answers in \acp{BBC}, e.g., greeting chitchats or opinions.

Therefore, most recent studies on \ac{BBC} focus on generation-based methods.
Since generation-based methods can generate natural and fluent responses, the key challenge is to find the appropriate background knowledge \citep{Lian2019Learning}.
\citet{zhang-2019-improving} introduce a pre-selection process that uses dynamic bi-directional attention to improve background \ac{KS} by using the utterance history context as prior information.
%
%
\citet{DBLP:conf/acl/LiNMFLZ19} devise an Incremental Transformer to encode multi-turn utterances along with background knowledge and design a two-pass decoder to improve \ac{KS}.
%
%
\citet{chuanmeng2019} combine the advantages of extraction-based and generation-based methods by incorporating a reference decoder that learns to select a span from the background during decoding.
\citet{liu2019knowledge} combine two types of knowledge, triples from knowledge graphs and texts from unstructured documents.
For \ac{KS}, they use multi-hop walking on graphs, like \citet{DBLP:conf/acl/MoonSKS19}.

Unlike the work described above, we address \ac{KS} in \acp{BBC} by introducing a novel \acf{GLKS} mechanism and a distantly supervised learning schema for better learning of the mechanism.
Most importantly, the proposed \ac{GLKS} needs neither span annotations \citep{chuanmeng2019} nor extra knowledge grounding \citep{liu2019knowledge}.


\section{\acl{GLKS}}

Given background material in the form of free text $K=[k_{1}$, $k_2, \ldots, k_t, \ldots, k_{|K|}]$, with $|K|$ tokens, and a current conversational context $X=[x_{1}, x_2, \ldots, x_t, \ldots, x_{|X|}]$, with $|X|$ tokens (usually, the previous $n$ utterances), the task of \ac{BBC} is to generate a response $Y=[y_{1}, y_2, \ldots, y_t, \ldots, y_{|Y|}]$ to $X$ by occasionally referencing background knowledge in $K$.

The proposed model \ac{GLKS}, shown in Figure~\ref{fig:02}, consists of four modules: \emph{Background \& Context Encoders}, a \emph{\acf{GKS} Module}, a \emph{State Tracker}, and a \emph{\acf{LKS} Module}.
Given $K$ and $X$, the background \& context encoders encode them into latent representations $\mathbf{H}^K$ and $\mathbf{H}^X$, respectively.
Then, the \ac{GKS} module evaluates the matching matrix between $\mathbf{H}^K$ and $\mathbf{H}^X$ globally.
Based on the matching matrix, \ac{GKS} makes a decision of ``what to talk about next'' by selecting continuous spans from the background $K$ to form a ``topic transition vector'' $\mathbf{h}_{X \rightarrow K}$.
At each decoding time step, \ac{LKS} outputs a response token by either generating from the vocabulary or selecting from the background $K$ under the guidance of the topic transition vector.

\subsection{Background and context encoders}
We use a bi-directional RNN with GRU \cite{cho-etal-2014-learning} to convert the background and context sequences into two hidden representation sequences $\mathbf{H}^K=$ $[\mathbf{h}^k_1$, $\mathbf{h}^k_2$, $\ldots$, $\mathbf{h}^k_t$, $\ldots$, $\mathbf{h}^k_{|K|}]$ and $\mathbf{H}^X=$ $[\mathbf{h}^x_1$, $\mathbf{h}^x_2$, $\ldots$, $\mathbf{h}^x_t$, $\ldots$, $\mathbf{h}^x_{|X|}]$, respectively:
\begin{equation}
\label{encoder}
\begin{split}
\mathbf{h}^k_t={}&\BiGRU^K(\mathbf{e}({k_t}), \mathbf{h}^k_{t-1}),
\end{split}
\end{equation}
where $\mathbf{e}(k_t)$ is the token embedding vector; $\mathbf{h}^k_0$ is initialized with 0;
$\mathbf{H}^X$ is obtained in a similar way but the $\BiGRU^X$ does not share parameters with $\BiGRU^K$.

\subsection{\acf{GKS} module}

Before calculating the match between $\mathbf{H}^K$ and $\mathbf{H}^X$, we first aggregate each representation in $\mathbf{H}^K$ and $\mathbf{H}^X$ with the last context output $\mathbf{h}^x_{|X|}$ using highway transformations \citep{Srivastava:2015:TVD:2969442.2969505}:
\begin{equation}
\label{ht}
\begin{split}
\mathbf{h}^k_t={}& g^k (\mathbf{W}_{linear}[\mathbf{h}^k_t, \mathbf{h}^x_{|X|}]+b)\\& + (1-g^k) \tanh(\mathbf{W}_{non\text{-}linear}[\mathbf{h}^k_t, \mathbf{h}^x_{|X|}]+b),\\
g^k={}&\sigma(\mathbf{W}_{gate}[\mathbf{h}^k_t, \mathbf{h}^x_{|X|}]+b]),
\end{split}
\end{equation}
where $\mathbf{W}_{linear}$, $\mathbf{W}_{non\text{-}linear}$ and $\mathbf{W}_{gate}$ are parameters; $b$ is a bias; and $\sigma$ is the sigmoid activation function.
We formulate background knowledge aggregation as above.
Context aggregation is achieved in a similar way.
Both aggregations can be performed multiple times so as to get deep representations.

Next, we estimate the transition matching matrix $\mathbf{M} \in \mathbb{R}^{|K|\times |X|}$ between $\mathbf{H}^K$ and $\mathbf{H}^X$, each element of which is calculated as follows:
\begin{equation}
\mathbf{M}[i,j] = \mathbf{v}_{M}^\mathrm{T}\tanh(\mathbf{W}_{M1}\mathbf{h}^k_i + \mathbf{W}_{M2}\mathbf{h}^x_j),
\end{equation}
where $\mathbf{v}_{M}$, $\mathbf{W}_{M1}$ and $\mathbf{W}_{M2}$ are parameters.
We apply max pooling along the $X$ dimension to get the transition weight vector $\mathbf{w}_{X \rightarrow K} \in \mathbb{R}^{|K|}$:
\begin{equation}
\mathbf{w}_{X \rightarrow K} = \max_{X} (\mathbf{M}).
\end{equation}
Each element of $\mathbf{w}_{X \rightarrow K}$ represents the transition possibility w.r.t. the corresponding token in $K$.

The weight vector $\mathbf{w}_{X \rightarrow K}$ only considers token-wise transitions.
However, a single token cannot determine the general meaning of the next response due to a lack of a global perspective.
To address this, we introduce the ``$m$-size unfold \& sum'' operation (as shown in Figure~\ref{fig:02}), which first extracts sliding adjacent weights of $\mathbf{w}_{X \rightarrow K}$ with an $m$-size window, and then sums them up.
Specifically, each element of the semantic unit transition weight vector $\mathbf{\hat{w}}_{X \rightarrow K}=[\mathbf{\hat{w}}_{X \rightarrow K_{0:m}}, \ldots, \mathbf{\hat{w}}_{X \rightarrow K_{L:L+m}}, \ldots]$ is calculated as follows:
\begin{equation}
\label{sut}
\mathbf{\hat{w}}_{X \rightarrow K_{L:L+m}} = \sum_{i=L}^{L+m} \mathbf{w}_{X \rightarrow K}[i].
\end{equation}
We assume there is no overlap between two adjacent semantic units, which helps to reduce the size of $\mathbf{\hat{w}}_{X \rightarrow K}$.

Correspondingly, we fuse the ``$m$-size unfold \& attention'' operation to obtain the semantic unit representations $\mathbf{\hat{H}}^K=[\mathbf{\hat{h}}_{K_{0:m}}, \ldots, \mathbf{\hat{h}}_{K_{L:L+m}}, \ldots]$ from $\mathbf{H}^K$:
\begin{equation}
\label{su}
\begin{split}
\mathbf{\hat{h}}_{K_{L:L+m}} ={}& \sum_{i=L}^{L+m} \alpha_i \mathbf{h}^k_i\\
\alpha_i ={}& \attention(\mathbf{h}^x_{|X|}, [\mathbf{h}^k_L, \ldots, \mathbf{h}^k_{L+m}]),
\end{split}
\end{equation}
where $\alpha_i$ is the additive attention weight between $\mathbf{h}^x_{|X|}$ and $\mathbf{h}^k_i$ \citep{attention2015}.
Note that $\alpha_i$ is normalized to probabilities with a local softmax operation (within the $m$-size window).
Each $\mathbf{\hat{h}}_{K_{L:L+m}}$ corresponds to a semantic unit (a text fragment) $K_{L:L+m}$ in background $K$.

Finally, we get the topic transition vector $\mathbf{h}_{X \rightarrow K}$ with a soft weighted average over $\mathbf{\hat{H}}^K$:
\begin{equation}
\begin{split}
\mathbf{h}_{X \rightarrow K} ={}& \sum_{L} P(K_{L:L+m}\mid X) \mathbf{\hat{h}}_{K_{L:L+m}} \\
P(K_{L:L+m}\mid X) \propto{}& \softmax(\mathbf{\hat{w}}_{X \rightarrow K_{L:L+m}}).
\end{split}
\end{equation}

\subsection{State tracker}

The state tracker is responsible for initializing the decoding state at the start and updating it at each following time step.
%
We get the initial decoding state $\mathbf{h}^s_{0}$ as follows:
\begin{equation}
\mathbf{h}^s_{0} = \mathbf{W}_s[\mathbf{h}^x_{|X|}, \mathbf{h}_{X \rightarrow K}]+b,
\end{equation}
where $\mathbf{W}_s$ is the parameter and $s$ is the bias.

For updating, we employ another GRU that takes the generated token and decoding state of the previous time step as input and outputs the updated decoding state:
\begin{equation}
\mathbf{h}^s_{t} =\GRU(\mathbf{e}(y_{t-1}), \mathbf{h}^s_{t-1}).
\end{equation}
Here, $y_0$ is set to a special token ``$<$BOS$>$,'' which indicates the start of decoding.

\subsection{\acf{LKS} module}
At each decoding time step, we use the \ac{LKS} module to predict each token one by one by either generating from vocabulary (with probability $P^V(y_t)$) or selecting from background $K$ (with probability $P^K(y_t)$) under the guidance of the topic transition vector $\mathbf{h}_{X \rightarrow K}$, as shown in Figure~\ref{fig:02}.

Specifically, we first concatenate $\mathbf{h}_{X \rightarrow K}$, $\mathbf{h}^s_{t}$ and $\mathbf{e}(y_{t-1})$ to get the guidance vector $\mathbf{h}^g_{t}$ at $t$:
\begin{equation}
\mathbf{h}^g_{t}=[\mathbf{h}_{X \rightarrow K}, \mathbf{h}^s_{t}, \mathbf{e}(y_{t-1})].
\end{equation}
Then, we employ background attention to get the guidance-aware background representation $\mathbf{\hat{h}}^K_{t}$ in Eq.~\ref{bcatt}:
\begin{equation}
\label{bcatt}
\begin{split}
\mathbf{\hat{h}}^K_{t} ={}& \sum_{i=1}^{|K|} \alpha^K_i \mathbf{h}^k_i,\\
\alpha^K_i ={}& \attention(\mathbf{h}^g_{t}, [\mathbf{h}^k_1, \ldots, \mathbf{h}^k_{|K|}]).
\end{split}
\end{equation}
In a similar way, we obtain the guidance-aware context representation $\mathbf{\hat{h}}^X_{t}$ with context attention.

We then construct the readout feature vector $\mathbf{\hat{h}}^r_{t}$ as follows:
\begin{equation}
\label{fv}
\mathbf{\hat{h}}^r_{t} = \mathbf{W}_r [\mathbf{e}(y_{t-1}), \mathbf{h}^s_{t}, \mathbf{h}_{X \rightarrow K}, \mathbf{\hat{h}}^K_{t}, \mathbf{\hat{h}}^X_{t}],
\end{equation}
where $\mathbf{W}_r$ are the parameter and $b$ is the bias.
The readout feature vector is then passed through a linear layer to estimate $P^V(y_t)$ with a softmax layer over the vocabulary:
\begin{equation}
P^V(y_t) = \softmax(\mathbf{W}_V \mathbf{\hat{h}}^r_{t}),
\end{equation}
where $\mathbf{W}_V \in \mathbb{R}^{|V|\times |F|}$ are the parameters, $|V|$ is the vocabulary size, and $|F|$ the hidden size of the readout feature vector $\mathbf{\hat{h}}^r_{t}$.

For $P^K(y_t)$, we employ another background attention as in Eq.~\ref{bcatt} to learn a pointer $\alpha^P_i$ as the probability of selecting a background token $k_i$.

Finally, we combine $P^V(y_t)$ and $P^K(y_t)$ as follows:
\begin{equation}
\begin{split}
P(y_t) ={}& g P^V(y_t) + (1-g) \sum_{y_t \in K} P^K(y_t) \\
g ={}& \sigma(\mathbf{W}\mathbf{h}^s_{t} + b),
\end{split}
\end{equation}
where $g$ is a learnable soft gate to switch between $P^V(y_t)$ and $P^K(y_t)$.

\subsection{Learning}
To maximize the prediction probability of the target response given the context and background, we design three objectives, namely the \acl{MLE} loss, the \acl{DS} loss, and the \acl{MCE} loss.

The \acfi{MLE} loss, which is commonly used, is defined as follows:
\begin{equation}
\begin{split}
\mathcal{L}_\mathit{mle}(\theta)=-\frac{1}{M}\sum_{m=1}^{M}\sum_{t=1}^{|Y|}\log P(y_t),
\end{split}
\end{equation}
where $\theta$ are all the parameters of our model, and $M$ is the number of training samples.

The \ac{MLE} loss only provides token-wise supervisions that lack a global perspective.
To address this, we define the \acfi{DS} loss to supervise the learning of \ac{GKS} (see Figure~\ref{fig:02}) as follows:
\begin{equation}
\begin{split}
\mathcal{L}_\mathit{ds}(\theta)={}&\frac{1}{M}\sum_{m=1}^{M} D_\mathit{KL} (P(\mathbf{\hat{H}}^K) \| Q(\mathbf{\hat{H}}^K))\\
P(\mathbf{\hat{H}}^K) ={}&\softmax(\mathbf{\hat{w}}_{X \rightarrow K})\\
Q(\mathbf{\hat{H}}^K) ={}&\softmax(\Jaccard(\hat{K}, Y)),
\end{split}
\end{equation}
where $\mathbf{\hat{w}}_{X \rightarrow K}$ is the semantic unit transition weight vector (Eq.~\ref{sut}) and $\mathbf{\hat{H}}^K$ are the semantic unit presentations (Eq.~\ref{su});
$Y$ is the ground truth response; $\hat{K}=[K_{0:m}$, \ldots, $K_{L:L+m}, \ldots]$ which is obtained with the same unfold operation as in Eq.~\ref{sut} or \ref{su}.
$D_\mathit{KL}$ is the KL-divergence, which is commonly used to measure the distance between two probability distributions;
$P(\mathbf{\hat{H}}^K)$ are the estimated probabilities of selecting the semantic units of $\mathbf{\hat{H}}^K$, which are obtained by using a softmax over the semantic unit transition weight vector; and, finally,
$Q(\mathbf{\hat{H}}^K)$ are the distant ground truth supervisions, which are obtained by calculating the Jaccard similarity between each semantic unit $K_{L:L+m}$ and the ground truth response $Y$.

Because $Q(\mathbf{\hat{H}}^K)$ is distance based, we use the \acfi{MCE} loss to alleviate the negative effects of the noise introduced by imprecise $Q(\mathbf{\hat{H}}^K)$:
\begin{equation}
\mbox{}\hspace*{-2mm}
\mathcal{L}_\mathit{mce}(\theta)\!=\!\frac{1}{M}\sum_{m=1}^{M}\sum_{t=0}^{|Y|}\sum_{w \in V}\!P(y_t\!=\!w) \log P(y_t\!=\!w).\!
\end{equation}
The final loss is a linear combination of the three loss functions:
\begin{equation}
\mathcal{L}(\theta)=\mathcal{L}_\mathit{mle}(\theta)+\mathcal{L}_\mathit{ds}(\theta)+\mathcal{L}_\mathit{mce}(\theta).
\end{equation}
All parameters of \ac{GLKS} are learned in an end-to-end back-propagation training paradigm.


\section{Experimental Setup}

\subsection{Implementation details}
For a fair comparison, we stay close to previous studies regarding hyper-parameters.
We set the word embedding size and hidden state size to 300 and 256, respectively.
The word embeddings are initialized with GloVe \citep{liu2019knowledge}.
The vocabulary size is limited to $\approx$26,000.
We limit the context length of all models to 65 \citep{moghe2018towards,chuanmeng2019}.
We select the best models of all methods according to the validation set.
We use gradient clipping with a maximum gradient norm of 2.
We use the Adam optimizer ($\alpha = 0.001$, $\beta_1 = 0.9$, $\beta_2 = 0.999$, and $\epsilon$ = $10^{-8}$).
We pre-train our model with the $\mathcal{L}_\mathit{ds}(\theta)$ loss for 10 epochs and then jointly train it with the other two losses.
The code is available online.\footnote{\url{https://github.com/PengjieRen/GLKS}}

\subsection{Dataset}
We choose the Holl-E dataset released by \citet{moghe2018towards} for experiments, because it is commonly used and contains the necessary information (boundary annotations, factoid knowledge) required by some recent methods~ \citep{chuanmeng2019,liu2019knowledge}.
It contains ground truth \ac{KS} labels that allow us to analyze the performance of models.
Holl-E is built for movie chats in which each response is explicitly generated by copying and/or modifying sentences from the background \citep{moghe2018towards}.
The background consists of plots, comments and reviews about movies collected from different websites.
Holl-E has three versions according to the background: oracle background (256-word), mixed-short background (256-word), and mixed-long background (1,200-word).
Oracle background has just one kind of background information (plots, comments, etc.).
We follow the original data split for training, validation and test, which contain 34,486, 4,388, and 4,318 samples respectively.
There are two versions of the test set: one with a single golden reference (SR), the other with multiple golden references (MR).

\subsection{Baseline}
We compare with all generation-based methods for which results on the Holl-E dataset have been reported at the time of writing:
\begin{itemize}[leftmargin=*,nosep]
\item \textbf{S2S} is a vanilla sequence-to-sequence model. 
\item \textbf{HRED} considers hierarchical modeling of context \citep{serban2016building}.
\item \textbf{S2SA} fuses an attention mechanism to do \ac{KS} at each decoding timestamp \citep{attention2015}. 
\item \textbf{GTTP} leverages a copying/pointer mechanism together with an attention mechanism to do \ac{KS} at each decoding timestamp \citep{see2017get}.
\item \textbf{Cake} introduces a pre-selection process that uses dynamic bi-directional attention to improve \ac{KS} \citep{zhang-2019-improving}.
\item \textbf{RefNet} combines the advantages of BiDAF \citep{seo2016bidirectional} and GTTP \citep{see2017get} by either selecting a span from the background with a reference decoder or generating a token with a generation decoder \citep{chuanmeng2019}.
\item \textbf{AKGCM} considers structured and unstructured knowledge for better \ac{KS} \citep{liu2019knowledge}. It uses policy network for \ac{KS} on structured knowledge and GTTP for \ac{KS} on unstructured knowledge and response generation.
\end{itemize}
S2S and HRED do not use any background knowledge; RefNet needs extra span annotations; AKGCM uses a structured knowledge graph and needs to manually ground knowledge between structured and unstructured sources.

\subsection{Evaluation metrics}
We use ROUGE-1, ROUGE-2 and ROUGE-L as automatic evaluation metrics.\footnote{We leave out BLEU since both previous and our experiments show that it has consistent performance with ROUGE~\citep{moghe2018towards,chuanmeng2019}}
Because the conversations are constrained by the background material, ROUGE scores are reliable.
Nevertheless, we also randomly sample 500 test samples to conduct human evaluations on Amazon Mechanical Turk.
For each sample, we show the responses from all systems to 3 workers and ask them to select all that are good\footnote{We allow for an ``all bad'' option.} in terms of four aspects: 
\begin{enumerate*}[label=(\arabic*)]
\item \emph{Naturalness} (\textbf{N}), i.e., whether the responses are conversational, natural and fluent; 
\item \emph{Informativeness} (\textbf{I}), i.e., whether the responses use some background information; 
\item \emph{Appropriateness} (\textbf{A}), i.e., whether the responses are appropriate/relevant to the given context; and 
\item \emph{Humanness} (\textbf{H}), i.e., whether the responses look like they are written by a human.
\end{enumerate*}


\section{Results}

\subsection{Automatic evaluation}
The results of all methods on different settings (oracle, mixed-short and mixed-long) are shown in Table~\ref{tab:result1}.

First, generally, \ac{GLKS} achieves the best results on all metrics.
\ac{GLKS} significantly outperforms two recent best performing methods (RefNet and AKGCM) on the mixed-short background.
The improvements show that \ac{GLKS} is much better at leveraging and locating the right background information despite  \ac{GLKS} not using any extra annotations (such as the span annotations used by RefNet) or information (such as the structured knowledge used by AKGCM).
We analyze the improvements of \ac{GLKS} in depth with an ablation study.

\begin{table}[t]
\caption{Automatic evaluation results (\%). \textbf{Bold face} indicates leading results in terms of the corresponding metric.
Significant improvements over RefNet are marked with $^\ast$ (t-test, p $<$ 0.01).
SR and MR refer to test sets with single and multiple references.
CaKe cannot run on the 1200-word background due to out of memory errors even with very small batch sizes \citep{zhang-2019-improving}.
The results of AKGCM are taken from the paper because the authors have not released their code.}
\label{tab:result1}
\centering
\resizebox{.95\columnwidth}{!}{\begin{tabular}{lcccccc}
\toprule
\multirow{2}{*}{} & \multicolumn{2}{c}{ROUGE-1} & \multicolumn{2}{c}{ROUGE-2} & \multicolumn{2}{c}{ROUGE-L} \\ \cmidrule(r){2-3}\cmidrule(r){4-5}\cmidrule(r){6-7}
                  & SR            & MR           & SR            & MR           & SR            & MR           \\ \midrule
\multicolumn{7}{c}{no background}                                                                            \\ \midrule
\small{S2S}               & 27.15         & 30.91        & 09.56         & 11.85        & 21.48         & 24.81        \\ 
\small{HRED}              & 24.55         & 25.38        & 07.61         & 08.35        & 18.87         & 19.67        \\ \midrule
\multicolumn{7}{c}{oracle background (256-word)}                                                                        \\ \midrule
\small{S2SA}              & 27.97         & 32.65        & 14.50         & 18.22        & 23.23         & 27.55        \\ 
\small{GTTP}              & 29.82         & 35.08        & 17.33         & 22.00        & 25.08         & 30.06        \\ 
\small{CaKe}              & 42.82         & 48.65        & 30.37         & 36.54        & 37.48         & 43.21        \\ 
\small{RefNet}            & 42.87         & 49.64        & 30.73         & 38.15        & 37.11         & 43.77        \\ 
\small{GLKS}              & \textbf{43.75}\rlap{$^\ast$}         & \textbf{50.67}\rlap{$^\ast$}        & \textbf{31.54}\rlap{$^\ast$}         & \textbf{39.20}\rlap{$^\ast$}        & \textbf{38.69}\rlap{$^\ast$}         & \textbf{45.64}\rlap{$^\ast$}        \\ \midrule
\multicolumn{7}{c}{mixed-short background (256-word)}                                                                           \\ \midrule
\small{S2SA}              & 26.36         & 30.76        & 13.36         & 16.69        & 21.96         & 25.99        \\ 
\small{GTTP}              & 30.77         & 36.06        & 18.72         & 23.70        & 25.67         & 30.69        \\ 
\small{CaKe}              & 41.26         & 45.81        & 29.43         & 34.00        & 36.01         & 40.79        \\ 
\small{RefNet}            & 41.33         & 47.00        & 31.08         & 36.50        & 36.17         & 41.72        \\ 
\small{AKGCM}             & --             & --            & 29.29         & --            & 34.72         & --            \\ 
\small{GLKS}              & \textbf{44.52}\rlap{$^\ast$}         & \textbf{50.06}\rlap{$^\ast$}        & \textbf{33.05}\rlap{$^\ast$}         & 38.87\rlap{$^\ast$}        & \textbf{39.63}\rlap{$^\ast$}         & \textbf{45.12}\rlap{$^\ast$}        \\ \midrule
\multicolumn{7}{c}{mixed-long background (1,200-word)}                                                                          \\ \midrule
\small{S2SA}              & 21.90         & 24.90        & 5.63          & 7.00         & 17.02         & 19.65        \\ 
\small{GTTP}              & 23.64         & 28.81        & 10.11         & 14.34        & 17.60         & 22.04        \\ 
\small{RefNet}            & 34.90         & 42.08        & \textbf{22.12}         & \textbf{29.74}        & 29.64         & 36.65        \\ 
\small{GLKS}              & \textbf{35.30}         & \textbf{42.31}        & 21.86         & 29.35        & \textbf{30.36}         & \textbf{37.30}        \\ \bottomrule
\end{tabular}}
\end{table}

Second, the improvements of \ac{GLKS} on the oracle and mixed-short background are much larger than on the mixed-long background.
The reason is that \ac{KS} becomes much more difficult when the background becomes longer.
This is supported by the fact that the results of all methods drop around 10\% compared with their results on the mixed-short background.
This also means that there is still a long way to go for \acp{BBC}.
\ac{GLKS} and RefNet are comparable in the mixed-long background setting.
\ac{GLKS} only gains around 0.3\% (ROUGE-1) and 0.7\% (ROUGE-L) improvement over RefNet.
RefNet is slightly better than \ac{GLKS} on ROUGE-2.
This is because RefNet uses extra span annotations, which shows great superiority in this setting.

\subsection{Human evaluation}

\begin{table}[t]
\caption{Human evaluation results. $\geq n$ means that at least $n$ MTurk workers think it is a good response w.r.t. \emph{Naturalness} (\textbf{N}), \emph{Informativeness} (\textbf{I}), \emph{Appropriateness} (\textbf{A}) and \emph{Humanness} (\textbf{H}).}
\label{tab:result2}
\centering
\resizebox{.95\columnwidth}{!}{\begin{tabular}{ccccccc}
\toprule
\multirow{2}{*}{} & \multicolumn{2}{c}{\textbf{Improved GTTP}} & \multicolumn{2}{c}{\textbf{RefNet}} & \multicolumn{2}{c}{\textbf{GLKS}} \\ 
\cmidrule(r){2-3} \cmidrule(r){4-5} \cmidrule{6-7} 
                  & $\geq$1                & $\geq$2                & $\geq$1                 & $\geq$2                & $\geq$1                & $\geq$2               \\ \midrule
\textbf{N}        & 307              & 115              & 391               & 213              & \textbf{424}     & \textbf{226}    \\ 
\textbf{I}        & 271              & \phantom{0}89              & \textbf{411}      & \textbf{244}     & 401              & 199             \\ 
\textbf{A}        & 318              & 111              & 371               & 180              & \textbf{406}     & \textbf{219}    \\ 
\textbf{H}        & 332              & 123              & 394               & 225              & \textbf{436}     & \textbf{263}    \\ \bottomrule
\end{tabular}}
\end{table}

We conduct human evaluations to further compare \ac{GLKS} and two strong baselines.
The results are shown in Table~\ref{tab:result2}.
The improved GTTP is equivalent to LKS in this paper. 
Both \ac{GLKS} and RefNet are better than GTTP on \emph{Naturalness} because GTTP frequently generates responses with no topics or irrelevant topics, which makes it difficult for mturk workers to assess the fluency.
RefNet gets the best votes on \emph{Informativeness} which means it invokes background knowledge more frequently.
This is consistent with its modeling schema, which encourages the model to refer to background during generation.
However, this does not mean \ac{GLKS} can always locate the appropriate background knowledge.
\ac{GLKS} achieves the best result on \emph{Appropriateness}, which means it is indeed better at \ac{KS} and can generate responses with more appropriate/relevant topics.
Unsurprisingly, \ac{GLKS} gets the most votes on \emph{Humanness} because its responses are more natural and appropriate.


\section{Analysis}

\subsection{Ablation study}

\begin{table}[t]
\caption{Ablation study (\%). 
-GKS, -$\mathcal{L}_{ds}(\theta)$ and -$\mathcal{L}_{mce}(\theta)$ denote \ac{GLKS} without the corresponding part.}
\label{tab:analysis1}
\centering
\resizebox{.95\columnwidth}{!}{\begin{tabular}{lcccccc}
\toprule
\multirow{2}{*}{} & \multicolumn{2}{c}{ROUGE-1} & \multicolumn{2}{c}{ROUGE-2} & \multicolumn{2}{c}{ROUGE-L} \\ \cmidrule(r){2-3}\cmidrule(r){4-5}\cmidrule(r){6-7}
                  & SR            & MR           & SR            & MR           & SR            & MR           \\ \midrule
-GKS          & 41.80         & 47.08        & 29.88         & 35.31        & 36.91         & 42.10        \\ 
-$\mathcal{L}_{ds}(\theta)$           & 41.27         & 46.96        & 29.49         & 35.40        & 36.47         & 42.12        \\ 
-$\mathcal{L}_{mce}(\theta)$           & 43.69         &  48.84       & 32.30         & 37.54        & 38.79         & 43.86        \\ 
GLKS              & \textbf{44.52}         & \textbf{50.06}        & \textbf{33.05}         & \textbf{38.87}        & \textbf{39.63}         & \textbf{45.12}        \\ \bottomrule
\end{tabular}}
\end{table}

To analyze where the improvements of \ac{GLKS} come from, we conduct an ablation study as shown in Table~\ref{tab:analysis1}.
Generally, all three parts (the GKS module, the \ac{DS} $\mathcal{L}_{ds}(\theta)$, and the \ac{MCE} $\mathcal{L}_{mce}(\theta)$) are helpful because removing any of them will decrease the results consistently.
GKS and $\mathcal{L}_{ds}(\theta)$ are much more effective because they yield around 3\% improvements.
This supports the motivations of our work which proposes to incorporate global perspective with distant supervision into \ac{KS}.
$\mathcal{L}_{mce}(\theta)$) is introduced to alleviate the negative effects of the noise introduced by imprecise distant supervisions.
The results of -$\mathcal{L}_{mce}(\theta)$ in Table~\ref{tab:analysis1} demonstrate its usefulness.
Even after removing all these modules, \ac{GLKS} still outperforms vanilla GTTP.
This is because we optimize the architecture with helpful tricks, e.g., using context states to aggregate background and context representations (like in Eq.~\ref{ht}), combining multiple representations to construct the readout feature vector (Eq.~\ref{fv}), etc.

\subsection{Hyper-parameter analysis}

\begin{figure}
\centering
\includegraphics[width=0.95\columnwidth]{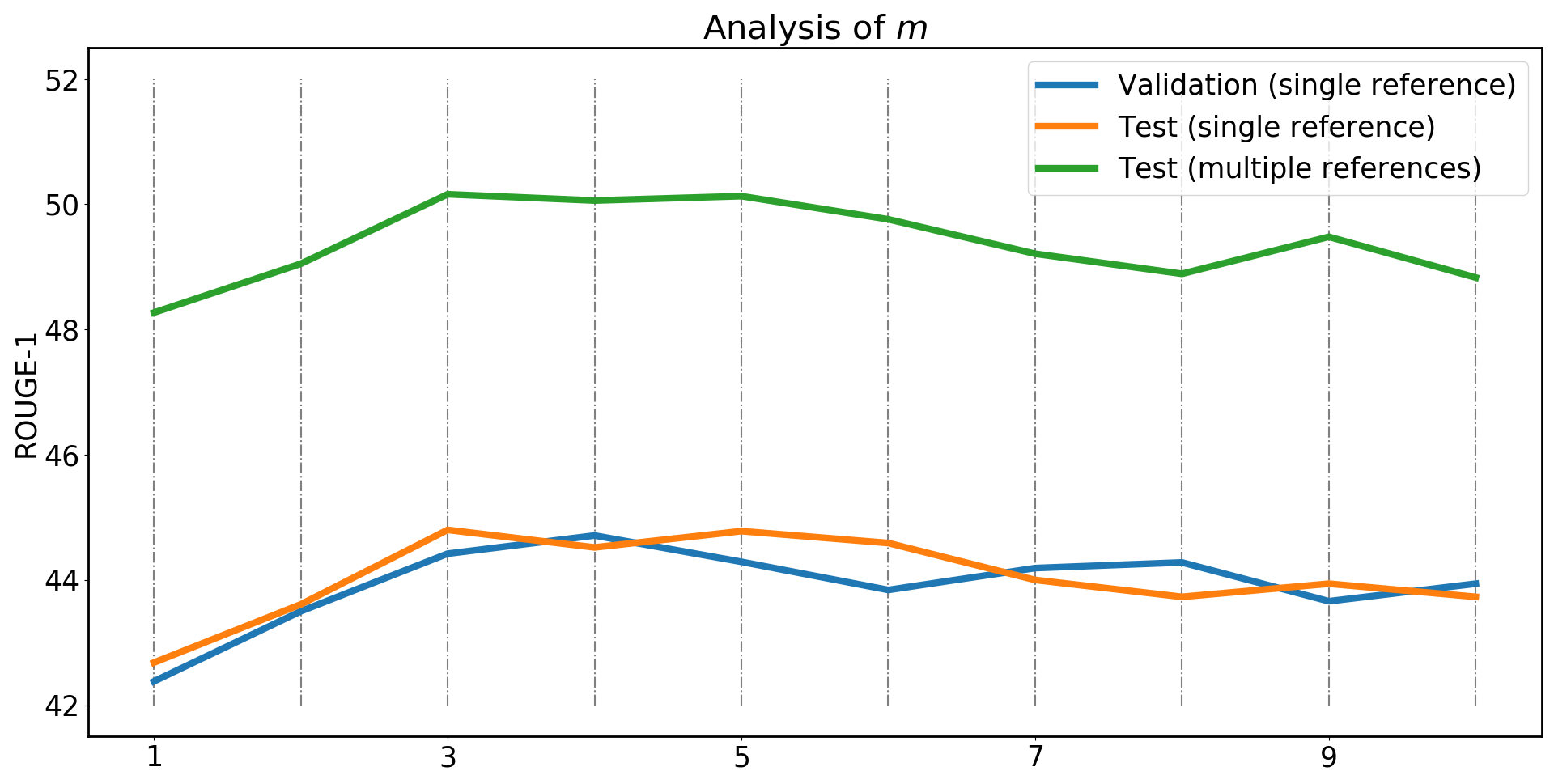} 
  \label{fig:03-1}
\caption{Analysis of $m$. The trends of ROUGE-2 and ROUGE-L are similar to ROUGE-1.}
\label{fig:03}
\end{figure}

There is a hyper-parameter $m$ that controls the unfolding window size in Eq.~\ref{sut} and \ref{su}.
We plot the ROUGE scores on the validation and test sets in Fig.~\ref{fig:03} to analyze its sensitivity.
The ROUGE scores increase and decrease within the scope of around 2\% difference which means \ac{GLKS} is not sensitive to $m$.
The best results are achieved around $m=3,4,5$ and the best validation results are achieved with $m=4$.
The results with $m \geq 3$ are much better than those with $m \leq 2$.
Hence, $m$ influences the performance and $m=4$ is enough to discriminate different knowledge and guide \ac{KS}.

\subsection{Visual analysis}

\begin{figure}
\centering
\begin{subfigure}{\columnwidth}
  \centering
  \includegraphics[width=0.95\columnwidth]{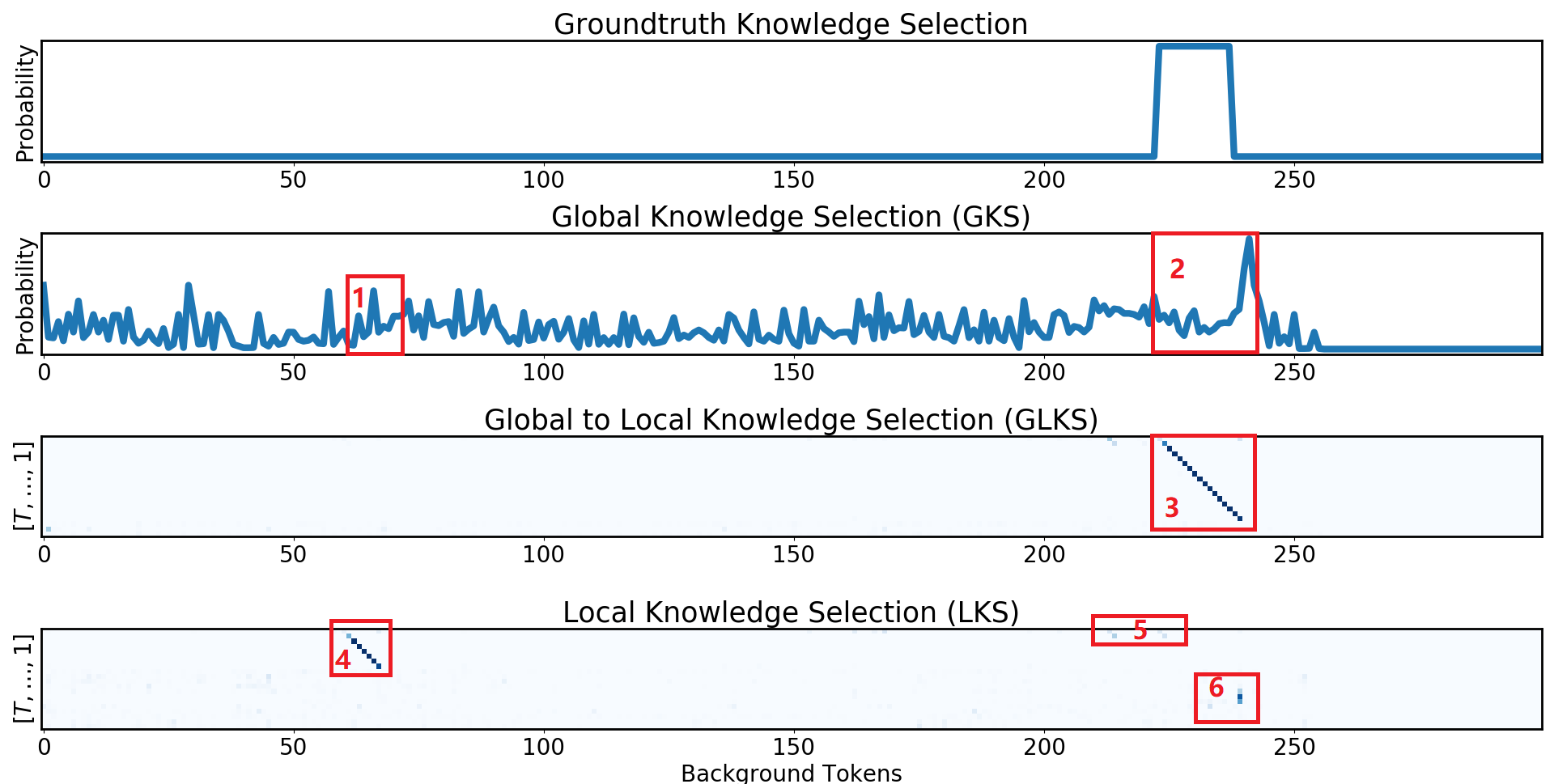}
  \caption{}
  \label{fig:04-1}
\end{subfigure}
\begin{subfigure}{\columnwidth}
  \centering
  \includegraphics[width=0.95\columnwidth]{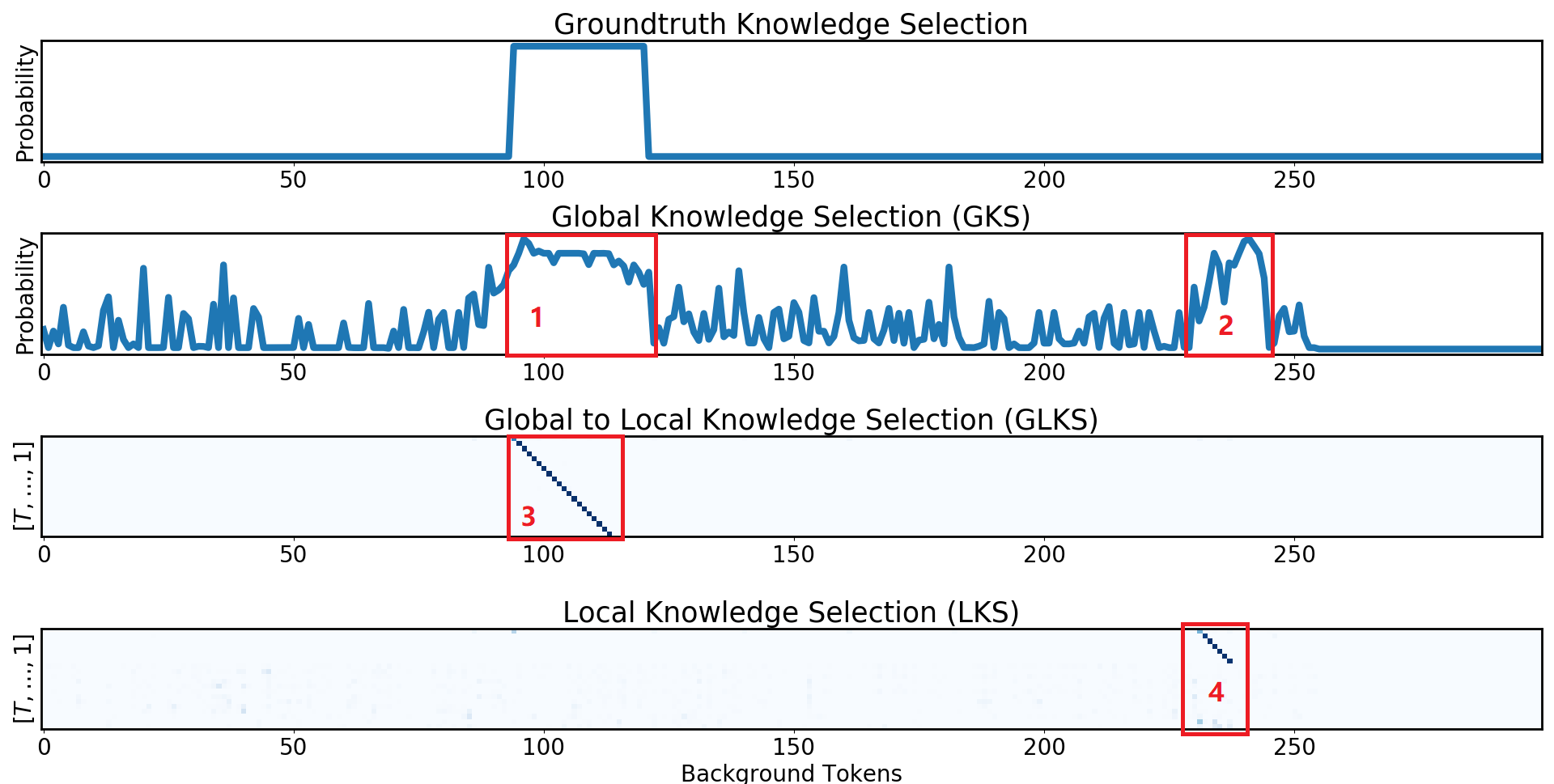}
  \caption{}
  \label{fig:04-2}
\end{subfigure}
\caption{\ac{KS} Visualization. For each figure, from top to bottom, are: ground truth \ac{KS}, \ac{GKS}, \ac{LKS} and \ac{GLKS}.}
\label{fig:04}
\end{figure}

In Fig.~\ref{fig:04} we visualize \ac{KS} with different settings.
The X axis corresponds to the background token sequence.
The Y axis of the first and left two figures denote \ac{KS} probabilities and decoding time steps, respectively.
The color depth in the lower two figure represents token-wise \ac{KS} probabilities.

We can see that without \ac{GKS}, \ac{LKS} can easily be fooled by similar but less appropriate knowledge (i.e., red box 1 and 2 in Figure~\ref{fig:04-1} and \ref{fig:04-2} respectively).
As a result, the model starts with the wrong or less appropriate knowledge (red box 4 in Figure~\ref{fig:04-1} and \ref{fig:04-2}) or results in inconsistent \ac{KS} (i.e., red box 4, 5, 6 in Figure~\ref{fig:04-1}) during generation.
This is because the model with only \ac{LKS} lacks a global perspective as guidance, making it harder to make decisions and easier to make mistakes.
In contrast, the model can avoid these issues and achieve better and more consistent \ac{KS} when taking \ac{GKS} into consideration (red box 3 in Fig.~\ref{fig:04-1} and \ref{fig:04-2}).

\subsection{Case study}

\begin{table}[hbt]
\caption{Case study.}
\label{tab:analysis2}
\centering
\scriptsize
\begin{tabular}{lm{7cm}}
\toprule
& 
 \textbf{Backgound}: ... {\color{red}later that evening , he intends to access kevin 's room , but kevin fools him into thinking that he has walked in on his father , causing the concierge to flee} ... home alone 2 is a carbon copy , but it 's also much better and more complex a movie than the first ... regardless it 's a classic and i watch the first two movies every year ... \\ \midrule
& \multicolumn{1}{l}{\begin{tabular}[l]{@{}p{6.8cm}@{}}\textbf{H1}: i loved all the tricks , and traps kevin created . \\ \textbf{H2}: me too , i loved when using a tape recorder , he tapes a message and slows down his voice , placing a hotel reservation . \\ \textbf{H1}: that was too funny , the hotel staff did n't believe him though .\end{tabular}}\\ \midrule
GTTP   & it 's a classic and i watch the first two movies every year . \\ \midrule
RefNet & {\color{red}that it was so sad when} he intends to access kevin 's room , but kevin fools him into thinking that he has walked in on his father , causing the concierge to flee . \\ \midrule
LKS    & i know , it was a carbon copy , but it 's also much better and more complex a movie than the first . \\ \midrule
GLKS   & {\color{red}so true ,} later that evening , he intends to access kevin 's room , but kevin fools him into thinking that he has walked in on his father , causing the concierge to flee . \\ \bottomrule
\end{tabular}
\end{table}

We select an example from the test set to intuitively illustrate the responses generated by different models, as shown in Table~\ref{tab:analysis2}.
We can see that all models have learnt to invoke knowledge during generation.
However, GTTP and LKS are relatively bad at \ac{KS}, resulting in using less appropriate knowledge.
RefNet is good at \ac{KS} and can generate natural responses.
But it has difficulties in coordinating the generation and reference decoding sometimes.
As a result, it has a higher probability of generating contradictory responses.
By comparison, \ac{GLKS} can generate appropriate responses which yields better humanness.

There are also failure cases for \ac{GLKS} as well as the other models: one severe issue is that the models tend to invoke the same knowledge even though the context has changed somewhat.
This indicates that a mechanism is needed to track the already used knowledge.


\section{Conclusion and Future Work}

In this paper, we propose an end-to-end neural model for \acp{BBC}, which introduces a \acf{GLKS} mechanism to enhance \ac{KS}.
We also present a \ac{DS} learning schema to learn \ac{GLKS} effectively without using any extra annotations or information.
Experiments show that with \ac{GLKS}, our model can generate more appropriate and human-like responses.

As to future work, we intend to apply \ac{GLKS} to other \ac{BBC} tasks.
Besides, \ac{GLKS} can be advanced in many directions.
First, better \ac{GKS} modules can be designed to further improve \ac{KS} especially when using very long background.
Second, a mechanism can be incorporated into \ac{GLKS} to enable the track of used knowledge in the context.

\section*{Acknowledgments}
We thank the anonymous reviewers for their helpful comments.
This work is supported by Ahold Delhaize, the Association of Universities in the Netherlands (VSNU), the Innovation Center for Artificial Intelligence (ICAI), the Natural Science Foundation of China (61672324, 61672322, 61972234, 61902219), the Natural Science Foundation of Shandong province (2016ZRE27468), the Tencent AI Lab Rhino-Bird Focused Research Program (JR201932), and the Fundamental Research Funds of Shandong University. All content represents the opinion of the authors, which is not necessarily shared or endorsed by their respective employers and/or sponsors.

\bibliography{bibtex}
\bibliographystyle{aaai}

\end{document}